## Brief Communications

# COVID-19 SignSym: a fast adaptation of a general clinical NLP tool to identify and normalize COVID-19 signs and symptoms to OMOP common data model

Jingqi Wang,[1,4,*] Noor Abu-el-Rub,[2,*] Josh Gray,[3] Huy Anh Pham,[1] Yujia Zhou,[4] Frank J. Manion,[1] Mei Liu,[2] Xing Song,[5] Hua Xu,[4] Masoud Rouhizadeh,[3,^] and Yaoyun Zhang[1,^]

[1]Melax Technologies, Inc, Houston, Texas, USA, [2]Division of Medical Informatics, University of Kansas Medical Center, Kansas City, Kansas, USA, [3]Johns Hopkins University School of Medicine, Baltimore, Maryland, USA, [4]School of Biomedical Informatics, The University of Texas Health Science Center at Houston, Houston, Texas, USA, and [5]University of Missouri School of Medicine, Columbia, Missouri, USA

*Contributed equally as first authors.
^Contributed equally as corresponding authors.

## ABSTRACT

The COVID-19 pandemic swept across the world rapidly, infecting millions of people. An efficient tool that can accurately recognize important clinical concepts of COVID-19 from free text in electronic health records (EHRs) will be valuable to accelerate COVID-19 clinical research. To this end, this study aims at adapting the existing CLAMP natural language processing tool to quickly build COVID-19 SignSym, which can extract COVID-19 signs/symptoms and their 8 attributes (body location, severity, temporal expression, subject, condition, uncertainty, negation, and course) from clinical text. The extracted information is also mapped to standard concepts in the Observational Medical Outcomes Partnership common data model. A hybrid approach of combining deep learning-based models, curated lexicons, and pattern-based rules was applied to quickly build the COVID-19 SignSym from CLAMP, with optimized performance. Our extensive evaluation using 3 external sites with clinical notes of COVID-19 patients, as well as the online medical dialogues of COVID-19, shows COVID-19 SignSym can achieve high performance across data sources. The workflow used for this study can be generalized to other use cases, where existing clinical natural language processing tools need to be customized for specific information needs within a short time. COVID-19 SignSym is freely accessible to the research community as a downloadable package (https://clamp.uth.edu/covid/nlp.php) and has been used by 16 healthcare organizations to support clinical research of COVID-19.

## INTRODUCTION

The COVID-19 pandemic[1] swept across the world rapidly resulting in the death, as of February 23, 2021, of at least 500 104 people in the US.[2] Scientists and researchers from multiple organizations worldwide have been working collaboratively, targeting effective prevention and treatment strategies.[3] Research findings, data resources, informatics tools, and technologies are being openly shared, aiming to speed up the fight against such emerging pandemics.[4,5]

Facilitated by the National Library of Medicine, large datasets of articles relevant to COVID-19 are being accumulated and shared at a rapid pace in the medical community.[6] For example, the COVID-19 Open Research Dataset has already accumulated more than 75 000 full text articles.[6] Based on such resources, many tools have been developed using natural language processing (NLP) techniques to unlock COVID-19 information from the literature, including search engines, information extraction, and knowledge graph building.[7,8] As another important data source for COVID-19 research, electronic health records (EHRs) store the clinical data of COVID-19 patients, which are critical for various applications, such as clinical decision support, predictive modeling, and EHR-based clinical research.[9,10] An efficient tool that can accurately recognize important clinical concepts of COVID-19 from clinical notes will be highly valuable to save time and accelerate the chart-review workflow.

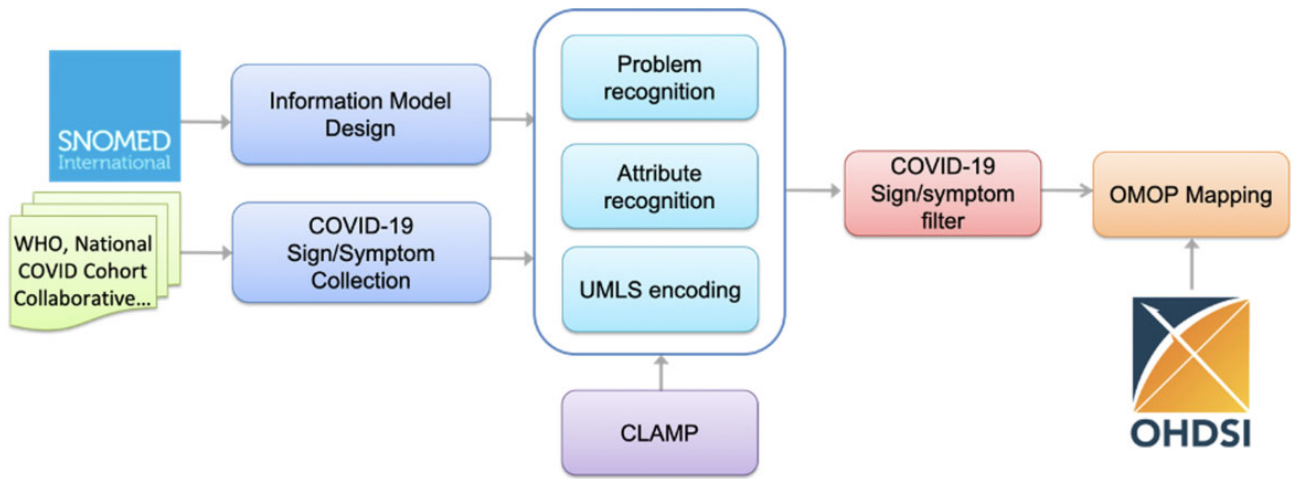

**Figure 1.** An overview of the NLP pipeline for COVID-19 sign/symptom extraction and normalization.

Despite that several large consortia have been formed to construct large clinical data networks for COVID-19 research, such as Open Health Data Science and Informatics (OHDSI),[11] the National COVID Cohort Collaborative (N3C),[12] and the international EHR-derived COVID-19 Clinical Course Profiles,[13] limited tools have been developed for COVID-19 information extraction from clinical notes. Rich patient information, such as signs/symptoms together with their attributes, is recorded in clinical narratives, which is critical for COVID-19 surveillance, diagnosis and prognosis, and treatment effectiveness research. Therefore, natural language processing (NLP) techniques are needed to automatically extract important clinical information efficiently and accurately from clinical text. Clinical NLP communities, such as the OHDSI NLP working group and the N3C NLP working group, have quickly reacted and built tools such as COVID-19 TestNorm[14] for lab testing normalization and MedTagger COVID-19 pipeline.[15,16]

Nevertheless, 1 common question worth answering is how to leverage existing general-purpose clinical NLP systems[17,18] to quickly respond to the urgent need of developing NLP pipelines for specific disease areas, such as COVID-19. Given that the datasets and annotations originally used to build the general-purpose clinical NLP systems are not dedicated to COVID-19 signs/symptoms, the learnt distributions of linguistic features and semantic types may not be sufficient to model the specific set of COVID-19 signs/symptoms, which leads to dropped performance for this specific task. To improve it, unique contextual patterns and lexicons may need to be created to tailor the existing clinical NLP system for COVID-19 information.

To address this question, in this study, we describe how to quickly build COVID-19 SignSym, which extracts COVID-19 signs/symptoms and 8 attributes (body location, severity, temporal expression, subject, condition, uncertainty, negation, and course) by adapting existing components in the Clinical Language Annotation, Modeling, and Processing (CLAMP) NLP tool.[17] The extracted entities were also normalized to standard terms in the OHDSI Observational Medical Outcomes Partnership (OMOP) common data model (CDM)[19] automatically. The tool was originally developed and evaluated on the MIMIC-III dataset. It was then evaluated using clinical notes of COVID-19 patients from 3 external institutions, including UT Physicians (UTP), University of Kansas Medical Center (KUMC), and Johns Hopkins Hospital. In addition, we also examined and adapted it to social media data—online medical dialogues of COVID-19. Evaluations on clinical notes and medical dialogues demonstrate promising results. We believe this tool will provide great support to the secondary use of EHRs to accelerate clinical research of COVID-19.

## MATERIALS AND METHODS

Figure 1 illustrates an overview of the workflow for building COVID-19 SignSym. This workflow mainly consisted of 5 steps: (1) Information model design to define the information scope of COVID-19 SignSym (ie, semantic types of clinical concepts and relations, to be extracted); (2) Sign/symptom collection from multiple sources to decide the scope of COVID-19 signs/symptoms; (3) Information extraction and normalization to extract and normalize COVID-19 signs/symptoms and 8 types of attributes, including body location, severity, temporal expression, subject, condition, uncertainty, negation, and course. Hybrid approaches were used to adapt existing CLAMP NLP pipelines for COVID-19 information extraction. Lexicons of COVID-19 signs/symptoms and pattern-based rules were integrated with CLAMP to optimize the performance. All signs/symptoms in free text were extracted first, then filtered by lexicons and Unified Medical Language System (UMLS) concept unique identifiers (CUIs) to obtain COVID-19-specific signs/symptoms. (4) OMOP mapping to convert the output information into the OMOP CDM concepts.

### Datasets
Five clinical datasets were used for building and evaluating the pipeline; these included MIMIC-III (used for pipeline construction and internal validation) and clinical notes data from UTP, KUMC, and

**Table 1.** A summary of statistics of each data source used for evaluation, including MIMIC-III, UTP, KUMC, Johns Hopkins, and medical dialogue

| Data Source | #notes/posts | #annotations of signs/symptoms | Sentences | Tokens |
| --- | --- | --- | --- | --- |
| MIMIC-III | 200 | 2214 | 17 208 | 329 044 |
| UTP | 100 | 3333 | 1357 | 5682 |
| KUMC | 100 | 527 | 4746 | 45 832 |
| Johns Hopkins | 334 | 467 | 13 397 | 121 802 |
| Medical dialogue | 100 | 602 | 1162 | 22 324 |

Johns Hopkins, as well as Medical Dialogs. The external clinical datasets were annotated independently at each site (ie, UTP, KUMC, and Johns Hopkins). Cohen's Kappa score was reported for interannotator agreement of each dataset between at least 2 annotators.

1. **MIMIC-III:** Given that no public EHR dataset of COVID-19 is available and that signs/symptoms of COVID-19 are a subset of general signs/symptoms, the MIMIC-III dataset was used for building the pipelines of information extraction and normalization in COVID-19 SignSym. MIMIC-III contains clinical notes from intensive patient care. 200 discharge summaries were randomly selected, with 50% for adapting the pipeline and 50% for open test. The reported interannotator agreement was 0.947. In total, the selected dataset contained 17 208 sentences and 329 044 tokens.

    The following datasets were used for external evaluation:

2. **Clinical notes from UTP:** UTP is the clinical practice of McGovern Medical School at The University of Texas Health Science Center at Houston. The dataset contained a sample of 20 patients who were tested for COVID-19 at UTP from 03/2020 to 06/2020. For those patients, 100 encounter notes were randomly selected. The gold standard labels were manually annotated for validating COVID-19 SignSym. Based on performance in an initial evaluation using a small set of clinical notes, 25 notes were used for tuning the pipeline, and 75 notes were used for open test. The reported interannotator agreement was 0.955. Overall, the annotated corpus contained 1357 sentences and 5682 tokens.

3. **Clinical notes from KUMC:** The dataset contained a sample of 20 hospitalized patients who were tested for COVID-19 and admitted to KU Hospital from 03/2020 to 08/2020. For those patients, 100 clinical notes were randomly selected from progress, history and physical examination (H&P), and emergency department (ED) notes, which were created 72 hours prior to or after hospital admission. All the notes were extracted from KUMC's EHR system (Epic), which has record boundaries limitations that could break 1 clinical note across multiple database records. Therefore, we performed a preprocessing step to concatenate broken records into complete notes and restored missing line breaks. The gold standard labels were manually annotated for validating COVID-19 SignSym. The interannotator agreement reported for the KUMC dataset is 0.892. Overall, the annotated corpus contained 4746 sentences and 45 832 tokens.

4. **Clinical notes from Johns Hopkins Hospital:** This dataset contained 334 clinical notes of 40 patients and includes relevant note types such as H&P, Critical Care Notes, Progress Notes, and ED Notes focusing specifically on the notes created within 48 hours before and after hospital admission.[20] Notes were preprocessed by Hopkins in-house section identification tools, and only the relevant narrative parts, particularly the chief complaint and history of the present illness sections, were extracted. For each of the 40 patients, each symptom was labeled as present or not present, resulting in over 467 manually annotated symptoms. These gold standard labels were then used to validate COVID-19 SignSym. The reported interannotator agreement was 0.970. In total, the clinical notes in this dataset contained 13 397 sentences and 121 802 tokens.

5. **Medical dialogues:** This dataset contained medical dialogues related to COVID-19 collected from an online website dialog between patients and doctors.[21] Initially, 50 dialogues were randomly selected for external evaluation. However, a sharp performance drop was observed on several attributes, such as temporal information, due to unique patterns in medical dialogues. Therefore, these 50 dialogues were used for tuning the pipeline and another 50 dialogues were used for an open test of COVID-19 SignSym. The reported interannotator agreement was 0.914. In total, the selected dataset contained 1162 sentences and 22 324 tokens.

The statistics of data sources used for evaluation in this study are summarized in Table 1.

### Information model

The information model followed by the COVID-19 SignSym is illustrated in Figure 2. In addition to mentions of signs and symptoms, 8 important attributes and their relations were recognized[22]: (1) Severity: indicates the severity degree of a sign/symptom; (2) Negation:

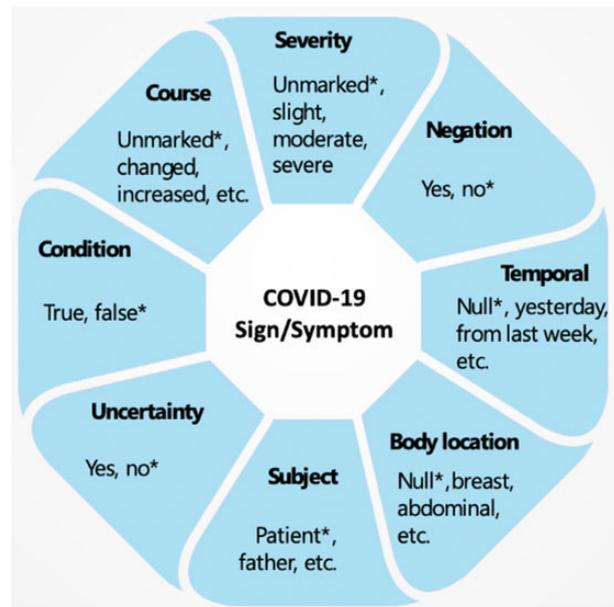

**Figure 2.** Information model of COVID-19 signs/symptoms and their attributes.

Table 2. Ten Examples of COVID-19 Signs and Symptoms, their synonyms, and UMLS CUIs

| Sign and Symptom | Example Synonym | UMLS CUI |
| --- | --- | --- |
| Sore throat | throat pain, throat soreness | C0242429 |
| Headache | head pain, cephalalgia | C0018681 |
| Fever | hyperthermia, febrile | C0015967 |
| Fatigue | tired, energy loss | C0015672 |
| Abdominal pain | stomach pain, gut pain | C0000737 |
| Altered consciousness | consciousness disturbances, impaired consciousness | C0234428 |
| Short of breath | sob, gasp, dyspnea | C0013404 |
| Dry cough | cough unproductive, nonproductive cough | C0850149 |
| Vomiting | throw up, puke | C0042963 |
| Diarrhea | loose stool, watery stool | C0011991 |

indicates a sign/symptom was negated; (3) Temporal information: indicates the time period or specific time the sign/symptom started; (4) Subject: indicates who experienced the sign/symptom; (5) Uncertainty: indicates a measure of doubt in a statement about a sign/symptom; (6) Condition: indicates conditional existence of sign/symptoms under certain circumstances; (7) Body location: represents an anatomical location of the sign/symptom; (8) Course: indicates progress or decline of a sign/symptom. Examples of attribute values are illustrated in Figure 2 (* indicates their default values).

### Lexicon building for COVID-19 signs and symptoms

Leveraging the community efforts, COVID-19 signs and symptoms were collected from 5 sources: (1) World Health Organization (WHO) case record form: 26 signs and symptoms were collected from the SIGNs AND SYMPTOMS ON ADMISSION section of the case record form provided by WHO.[23] (2) N3C: 15 signs and symptoms were collected from the diagnosis table shared by the N3C as phenotyping information.[12] (3) MedTagger Lexicon: 17 signs and symptoms together with their 136 synonyms were collected from the lexicon in MedTagger.[16] (4) Lexicon from Johns Hopkins: 14 signs and symptoms, and 337 synonyms were collected from an in-house lexicon of Johns Hopkins. (5) UMLS CUIs of these signs and symptoms were also assigned manually, with their UMLS synonyms collected. After removing redundancy, 55 signs and symptoms (with 153 different CUIs in total) and 3213 synonyms were collected from these sources. Notably, the reason why 55 signs and symptoms have 153 UMLS CUIs is that some signs/symptoms are represented by multiple semantic concepts with finer-granularities in UMLS. For example, productive cough is represented as productive cough—green sputum (CUI: C0425504), productive cough—clear sputum (CUI: C0425505), and productive cough—yellow sputum (CUI: C0425506) in UMLS.

Table 2 illustrates 10 signs/symptoms of COVID-19 together with their CUIs and example synonyms. A comprehensive list of COVID-19 signs, symptoms, and their corresponding UMLS CUIs and synonyms can be accessed through GitHub.[24]

### Building pipelines of information extraction and normalization using CLAMP

**Problem-attributes pipeline in CLAMP**: The problem-attribute pipeline in CLAMP[18] was adapted for COVID-19 information extraction in this study. This pipeline was built to automatically extract mentions of problems and their 8 attributes from clinical text. The definition of problems follows that used in the i2b2 2010 shared task, which consisted of 11 semantic types in UMLS (eg, sign or symptom, pathologic functions, disease or syndrome, etc).[25] The definitions of attributes follow those used in the SemEval 2015 shared task 14.[22]

The pipeline used a hybrid approach to recognize mentions of problems and their attributes by combining the deep learning-based method, domain lexicons, and rules. Both the recognition of problems and attributes were represented as a sequential labeling task.[26] In particular, a novel joint-learning framework was designed to recognize attributes and their relations with problems simultaneously by integrating problem mentions as features into the deep learning architecture.[27] To balance the time efficiency and performance in a practical clinical NLP tool[28], the deep learning-based Bi-directional Long Short-Term Memory Conditional Random Fields (BiLSTM-CRF) algorithm[26] was used to recognize mentions of clinical concepts, with 200-dimension word embeddings of MIMIC-III as the initial input.

**Dictionary lookup for COVID-19 signs and symptoms**: The lexicons previously collected were used in an additional step of dictionary-lookup to improve the coverage of recognized COVID-19 signs and symptoms. Regular expressions and rules were also applied in a postprocessing step to boost the performance of attribute recognition.

**Concept normalization**: Mentions of problems and attributes were normalized to standard concepts using the UMLS encoder module of CLAMP, which were built on a hybrid method of semantic similarity-based ranking and rules of concept prevalence in UMLS.[29] Both CUIs and preferred terms in UMLS were output for each recognized entity.

**Filtering of COVID-19 signs/symptoms**: Once medical problems were automatically recognized and normalized to UMLS CUIs, they were filtered by the precollected lexicons and CUIs of COVID-19 signs and symptoms.

**OMOP concept mapping**: The remaining signs/symptoms and their attributes were also mapped to standard concepts in the OMOP CDM. The OMOP encoder module of CLAMP was used for this purpose, which applied a similar approach as in the UMLS encoder module with a different scope of standard concepts and identifiers. Finally, COVID-19 SignSym outputs tab-delaminated files of normalized clinical concepts, with a format ready for importing into the Note_NLP table of OMOP CDM.[30]

### Evaluation

**Evaluation of information extraction**: (1) *Internal evaluation on MIMIC-III*: As mentioned above, COVID-19 SignSym is a pipeline adapted from existing clinical information extraction pipelines in CLAMP. To tune this pipeline, 200 discharge summaries were ran-domly selected from MIMIC-III based on lexicons of COVID-19 signs/symptoms; 100 of them were used for error analysis and optimizing the SignSym pipeline; after that, the information extraction performance was evaluated on another 100 discharge summaries as the open test. (2) *External evaluation on COVID-19 data*: the tool was evaluated on 100 clinical notes of patients tested for COVID-19 in UTP. In addition, it was evaluated on 100 posts of COVID-19-related dialogues between patients and doctors online as an external test. Notably, an initial evaluation was carried out on each external source with a small set of data to determine if the SignSym pipeline needs to be fine-tuned to adapt to the external data for better performance. If adaptation was needed, the initial data was used for error analysis, so that common patterns in each data source could be summarized for performance improvement. As a result, 25 clinical notes from UTP were used for fine-tuning the pipeline and 75 were used for open test. In addition, 50 medical dialogues were used for fine-tuning the pipeline and another 50 were used for open test.

Errors present in the outputs of COVID-19 SignSym are analyzed carefully for adapting the pipeline to external datasets. As illustrated in Supplementary Table 1, common causes of errors can be categorized into several major types:

For UTP notes: (1) False positive signs and symptoms within wrong context: since a dictionary lookup method using sign/symptom lexicons of COVID-19 is carried out to improve the recall, some lexicons are matched in a wrong context, such as flu in "flu shot" and influenza in "Influenza vaccine." Semantic types in UMLS were used to filter out such false positives. (2) Semistructure list: some attributes, such as conditions, are modifying signs/symptoms in a list of multiple items. Regular expressions were designed to parse common semistructure patterns.

For medical dialogues: (3) New temporal patterns: some temporal expressions have new patterns and contexts uncovered in previous training datasets, such as in "Saturday February 29th—diarrhea." To address this problem, new temporal terms and pattern-based rules were added to the pipeline.

For both UTP notes and medical dialogues: (4) New synonyms of signs/symptoms: for example, "breathing harder," "cannot breathe" (original typos are kept here) are mentions of "shortness of breath." Another error is caused by partial recognition of named entities: For example, "cough" is recognized instead of "cough productive of yellow sputum." Unseen mentions were used to enrich the lexicon to improve its coverage. (5) Question marks as false positive uncertainty: both clinical text and medical dialogues contain questions about patients' signs and symptoms. According to our current annotation guideline, if the question has answers along with it, then the question mark will not be considered as a mention of uncertainty; or else, it will be labeled as an uncertainty attribute. (6) New patterns of subjects: many subjects in medical dialogues are family members of the question asker, different from the common cases in clinical text, such as in "my daughter is 3 and a half… has had pneumonia." Besides, false positives also occur when the family member's statement is cited in the clinical text, such as in "Per his wife he had an episode of confusion." Since the subject attribute is of low frequency in the text, it is hard to capture the complex context of it. Syntactic structures will be applied in the next step in attempt to resolve this problem.

**Evaluation of concept normalization**: 1000 terms were randomly selected from the UTP notes and manually reviewed to evaluate the performance of clinical concept normalization.

**Use case-based evaluation**: A use case of identifying patients' signs/symptoms at presentation from notes generated within 72/48 hours before and after admission at KUMC/Johns Hopkins Hospital was used to validate the effectiveness of COVID-19 SignSym. Specifically, the task was to recognize signs and symptoms of the patient, in order to assist in future COVID-19 diagnosis and research. Specifically, once signs and symptoms were recognized in clinical text and normalized, they were also determined as negated, possible, or positive using attributes in their context. Only positive signs/symptoms were kept, and they were aggregated into the patient level to compare with the gold standard annotations of signs and symptoms manually recorded for each patient.

**Evaluation criteria**: (1) The end-to-end performances of extracting signs, symptoms, and their attributes were evaluated using precision, recall, and F-measure (F1) by exact match between manual and automatic annotations. (2) The performance of concept normalization was evaluated using accuracy. (3) The performances of patient-level extraction of COVID-19 signs/symptoms were evaluated using precision, recall, and F-measure.

## RESULTS

**Information extraction**: Internal evaluation performances for COVID-19 information extraction on MIMIC-III are reported in Table 3. The F-measure of signs/symptoms was enhanced from 0.846 to 0.972 after improving the original CLAMP pipeline for COVID-19 information. Especially, the recall was increased by 27.02%—from 0.781 to 0.992. Given that a joint learning approach was applied for recognizing the mentions of attributes and their relations with signs/symptoms, the performance enhancement on signs/symptoms was also propagated to the attribute recognition. The recalls of attributes also increased sharply. For example, the recall of body location was improved from 0.551 to 0.965, leading to an improvement of the F-measure from 0.688 to 0.954. For attributes with low frequencies, their performances are relatively modest. For example, severity has only 24 instances in the test dataset, with an F-measure of 0.857.

External evaluation performances on clinical notes of UTP and the medical dialogue are illustrated in Table 4. The performances of the original COVID-19 SignSym pipeline and after adaptation are reported. As shown in Table 3, performances of COVID-19 SignSym are improved after adapting the pipeline to external data, especially for the 3 attributes—temporal expression, condition, and uncertainty. Notably, promising results were achieved on sign/symptom extraction, with F-measures of 0.986 on UTP notes and 0.99 on the medical dialogues. As for recognizing attributes of signs/symptoms, the tool yielded better performances on attributes with higher frequencies (eg, F-measure: Body location, 0.992 for UTP notes and 0.94 for medical dialogue; Negation, 0.999 for UTP notes and 1.00 for medical dialogue). Similar to the internal evaluation, some semantic types have low frequencies in the test datasets with modest performances. For example, there are only 7 mentions of Subject (F-measure: 0.842) and 16 mentions of Uncertainty (F-measure: 0.741) in clinical notes of UTP.

**Table 3.** Information extraction performances on clinical text of MIMIC-III for internal evaluation, using the original CLAMP pipeline and COVID-19 Sign/Sym, respectively

| Data Source | MIMIC-III | | | | | |
|---|---|---|---|---|---|---|
| Pipeline | Original CLAMP Pipeline | | | COVID-19 Sign/Sym | | |
| Criteria | P | R | F1 | P | R | F1 |
| Sign/Sym | 0.92 | 0.781 | 0.846 | 0.953 | 0.992 | 0.972 |
| Attribute Recognition | | | | | | |
| Body location | 0.914 | 0.551 | 0.688 | 0.938 | 0.965 | 0.954 |
| Temporal | 0.85 | 0.354 | 0.496 | 0.922 | 0.979 | 0.95 |
| Negation | 0.984 | 0.375 | 0.543 | 0.994 | 0.994 | 0.994 |
| Condition | 0.959 | 0.723 | 0.824 | 1 | 1 | 1 |
| Course | 0.97 | 0.444 | 0.609 | 0.984 | 0.861 | 0.918 |
| Uncertainty | 0.938 | 0.577 | 0.714 | 0.85 | 0.981 | 0.911 |
| Severity | 0.5 | 0.5 | 0.5 | 1 | 0.75 | 0.857 |
| Subject | 1 | 0.667 | 0.8 | 1 | 1 | 1 |

**Table 4.** Information extraction performances of COVID-19 SignSym on clinical text and medical dialogue for external evaluation

| Data Source | Clinical Text in UTP | | | | | | Medical Dialogue | | | | | |
|---|---|---|---|---|---|---|---|---|---|---|---|---|
| Pipeline | COVID-19 SignSym | | | With adaptation | | | COVID-19 SignSym | | | With adaptation | | |
| Criteria | P | R | F1 | P | R | F1 | P | R | F1 | P | R | F1 |
| Sign/Sym | 0.978 | 0.82 | 0.947 | 0.96 | 0.984 | 0.986 | 0.918 | 0.915 | 0.961 | 0.98 | 0.97 | 0.99 |
| Attribute recognition | | | | | | | | | | | | |
| Body location | 0.965 | 0.797 | 0.937 | 0.971 | 0.994 | 0.992 | 0.933 | 0.757 | 0.836 | 0.971 | 0.92 | 0.94 |
| Temporal | 0.838 | 0.823 | 0.92 | 0.849 | 0.94 | 0.952 | 0.833 | 0.556 | 0.667 | 0.853 | 0.81 | 0.83 |
| Negation | 0.997 | 0.834 | 0.955 | 0.996 | 0.995 | 0.999 | 0.882 | 0.75 | 0.811 | 1.00 | 1.00 | 1.00 |
| Condition | 0.879 | 0.879 | 0.879 | 0.861 | 0.946 | 0.961 | 1.00 | 0.65 | 0.788 | 1.00 | 1.00 | 1.00 |
| Course | 1.00 | 0.94 | 0.969 | 1.00 | 0.964 | 0.982 | 11.00 | 0.94 | 0.969 | 1.00 | 1.00 | 1.00 |
| Uncertainty | 0.75 | 0.375 | 0.500 | 0.909 | 0.625 | 0.741 | 0.75 | 0.375 | 0.50 | 0.781 | 0.89 | 0.83 |
| Severity | 1.00 | 0.75 | 0.857 | 1.00 | 1.00 | 1.00 | 0.875 | 0.778 | 0.824 | 1.00 | 0.75 | 0.85 |
| Subject | 0.25 | 1.00 | 0.40 | 0.70 | 1.00 | 0.824 | 1.00 | 0.45 | 0.621 | 1.00 | 0.70 | 0.82 |

**Concept normalization:** Based on manual check of 1000 gold standard entities annotated in the UTP notes and their automatically assigned OMOP concepts, COVID-19 SignSym obtained an accuracy of 91.8% for concept normalization.

**Use case-based evaluation:** (1) COVID-19 sign/symptom presentation of patients at KUMC: In comparison with manually assigned signs/symptoms of each patient, the tool yielded a precision of 0.908, a recall of 0.917, and an F-measure of 0.909. (2) COVID-19 sign/symptom presentation of patients at Johns Hopkins Hospital: In comparison with manually assigned signs/symptoms of each patient, the tool yielded a precision of 0.928, a recall of 0.957, and an F-measure of 0.942.

### Tool availability

This tool is freely accessible to the research community as a downloadable package (https://clamp.uth.edu/covid/nlp.php).[18] A visualization of the pipeline output is illustrated in Figure 3.

## DISCUSSION

There is an urgent demand for automated tools that can extract COVID-19 information from clinical text. This article presents our attempt to quickly adapt an existing general clinical NLP tool for COVID-19 sign/symptom extraction and normalization, leveraging COVID-19 lexicons provided from multiple resources. The first version of the COVID-19 SignSym was released on May 10, 2020, probably 1 of the earliest NLP tools for COVID-19 information extraction from clinical notes. Since then, over 16 healthcare organizations have used the tool to support diverse applications including those in KU and Johns Hopkins Hospital, indicating the utility of such a tool in the pandemic.

Several interesting findings have been observed, which may be generalized to other use cases, where existing clinical NLP tools need to be quickly customized for specific information needs. At the beginning of the pandemic, little COVID-19 clinical data (eg, notes) were available for developing data-driven approaches (eg, machine learning-based ones). Therefore, rule-based approaches that leverage knowledge from domain experts or existing knowledge bases could be built quickly and our results also show that such methods could achieve good performance on sign and symptom recognition. Moreover, our results also show that previously existing clinical corpora of similar diseases (eg, signs and symptoms of influenza in MIMIC-III) could be useful for developing methods for the new disease. Additionally, the deep learning-based disease-modifier models in the CLAMP tool also demonstrated reasonable performance when they were applied to COVID-19 signs and symptoms, indicating the adaption process could be mainly focused on defining postprocessing rules (instead of reannotating text and retraining models), thus saving the development time for the new NLP pipelines.

Nevertheless, the developed COVID-19 SignSym pipeline did not always perform well when applied to different sites. For example, it worked well on KU and Johns Hopkins corpora; but direct application to UTP notes showed a drop in performance and required local customization, especially for certain types of entities and probably due to the intrinsic syntactic and semantic variance of clinical notes from different sites. As indicated from the error analysis, attributes with low frequencies in medical text suffer from a relatively sharper performance drop when applying the pipeline to a new dataset, probably because of new patterns uncovered in the previous training data. Also, the low frequencies make their performances vulnerable. For example, if an attribute has only 20 occurrences in the dataset, 1 false positive will yield a 5% drop in precision. Despite using pretrained language models of clinical text and deep learning-based methods for model building, there is still space for improvement, such as increasing the scale and coverage of the unlabeled clinical text used to build the pretrained language model and automatically adapting the current pipelines to new datasets.

It would be interesting to investigate the lexicon variations of COVID-19 signs/symptoms among multiple sites. After removing lexical redundancies, the overlap among signs/symptoms annotated on MIMIC-III, UTP notes, and medical dialogues are illustrated in the Venn diagram of Figure 4. Only 31 mentions of signs/symptoms (eg, abdominal pain, cough, chest pain) are present in all 3 data sources, while 23.2% (111/478) in UPT, 33.5% (73/215) in medical dialogs and 19.8% (98/496) in MIMIC-III are overlapped with other data sources. One potential reason for this large lexicon variation is that the data sources are of different settings, each of their unique sublanguages: MIMIC-III are mainly discharge summaries from ICU, UTP notes are encounter notes of potential COVID-19 patients, and medical dialogues are mainly written and posted by consumers. For example, cold was mentioned as a bad cold, a cold, and a head cold in medical dialogue, while none of these mentions present in the other 2 data sources. Despite of such large lexicon variations, the proposed hybrid method of combining deep learning models, dictionary lookup and pattern-based rules was able to catch the majority of mentions of COVID-19 signs/symptoms (F-measure: 0.972 MIMIC-III, F-measure: 0.986 UTP, F-measure: 0.99 medical dialogue), indicating the feasibility of quickly adapting general clinical NLP systems for information extraction of specific disorders.

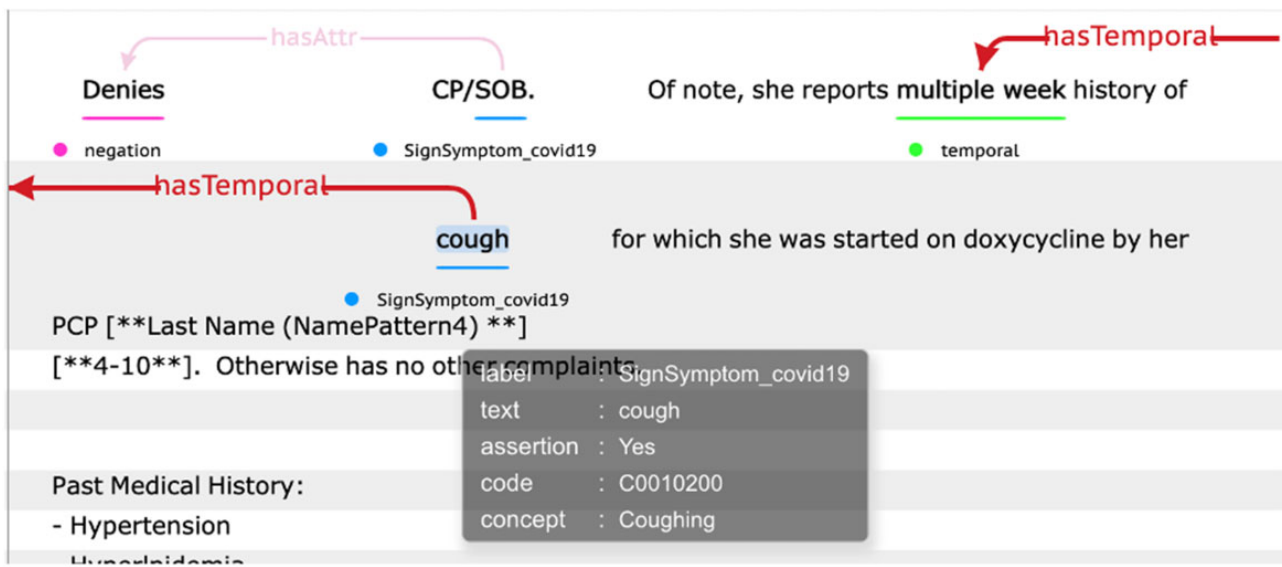

**Figure 3.** An output illustration of the COVID-19 sign/symptom extraction tool.

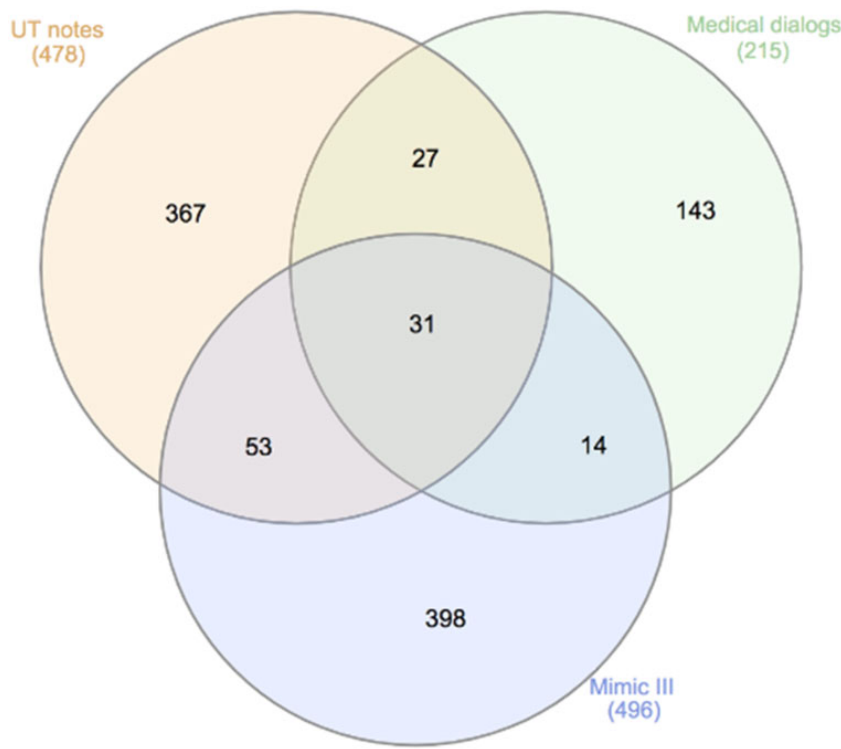

**Figure 4.** A Venn diagram of lexicon overlap among MIMIC-III, UTP notes, and medical dialogues for COVID-19 signs/symptoms.

Limitations and future work: this work has several limitations and future work is needed in several areas. While performance on 5 COVID-19 datasets is evaluated and reported, additional evaluations are needed to further refine the tool and increase its generalizability. At this stage, lexicon and pattern-based rules are used for a fast adaptation of the pipeline to new datasets, based on detailed er-ror analysis. This will serve as a foundation for later automatic adaptation such as semisupervised learning and automatic pattern detection when external datasets of larger sizes are available. Note that extracted phenotyping information representations will be applied in downstream clinical applications in our collaborator's site. Currently, only sign/symptoms and their attributes are extracted; additional work will be conducted for more related information such as comorbidities and medications. Finally, the output information will also be mapped to other clinical data standards, such as FHIR, in the near future to facilitate clinical operations and other applications.

## CONCLUSION

This article presents an automatic tool, COVID-19 SignSym, to extract and normalize signs/symptoms and their 8 attributes from clinical text by adapting a general clinical NLP tool (CLAMP) to COVID-19 information. COVID-19 SignSym achieved great performance in external evaluation using clinical notes of COVID-19 patients from 3 institutions and a social media dataset of medical dialogues and has been used to facilitate COVID-19 research. The workflow used for this study may be generalized to other use cases, where existing clinical NLP tools need to be customized for specific information needs within a short time.

## FUNDING

This study is partially supported by the National Center for Advancing Translational Sciences grant number R44TR003254 and CTSA grant number UL1TR002366.

## AUTHOR CONTRIBUTIONS

The work presented here was carried out in collaboration among all authors. YZ, MR, and HX designed methods and experiments. NA, MR, JG, HP, YZ, XS, ML, and JW annotated the datasets. HP, JW, NA, and MR analyzed the data and interpreted the results. YZ and FM drafted the article. All authors have contributed to, edited, reviewed, and approved the manuscript.

## DATA AVAILABILITY STATEMENT

Annotation datasets of MIMIC-III and medical dialogues are available on request.

## SUPPLEMENTARY MATERIAL

Supplementary material is available at *Journal of the American Medical Informatics Association* online.

## CONFLICT OF INTEREST STATEMENT

Dr Hua Xu, Mr Jingqi Wang, and The University of Texas Health Science Center at Houston have financial related research interest in Melax Technologies, Inc.